\pdfoutput=1
\documentclass[letterpaper, 10 pt, conference]{ieeeconf}
\IEEEoverridecommandlockouts
\usepackage{cite}
\usepackage{amsmath,amssymb,amsfonts}
\usepackage{graphicx}
\usepackage{textcomp}
\usepackage{xcolor}
\usepackage{booktabs}
\usepackage{url}

\title{Training-Independent Validation of a Learning-Based \\Controller for Stunt Maneuvers}

\author{Kiruthiga Chandra Shekar$^{1}$, Aliasghar Arab$^{2}$, and Jingang Yi$^{3}$%
\thanks{$^{1}$K.\ C.\ Shekar is with General Autonomy, USA.
        {\tt\small kshekar@genauto.ai}.}%
\thanks{$^{2}$A.\ Arab is with the City College of New York, New York University, and General Autonomy, USA.
        {\tt\small aarab@ccny.cuny.edu, aliasghar.arab@nyu.edu, aarab@genauto.ai}.}%
\thanks{$^{3}$J.\ Yi is with the Department of Mechanical and Aerospace Engineering, Rutgers University, Piscataway, NJ, USA.
        {\tt\small jgyi@rutgers.edu}.}%
}

\begin{document}

\maketitle
\thispagestyle{empty}
\pagestyle{empty}

\begin{abstract}
Learned controllers for vehicle maneuvers are usually validated under the conditions and objective used to train them, where they appear highly reliable, yet the parameters that change on real roads are often not those the model was trained on. We argue that validation must be independent of training: it should vary physical parameters that training kept fixed and use acceptance criteria absent from the training objective. A learning-based control policy, G-Maneuver, with 90.11\% success is tested for stunt-level maneuvers. A looser terminal-heading score ranked the worst setting as the best. Because real stunt tests are safety-critical, a simulation-to-reality-to-simulation (Sim2Real2Sim) loop uses the simulation to select safe zones for real-vehicle trials with 16.9\% acceptance. Automotive safety analysis frameworks are used to identify parameters for validation using coverage tests. We demonstrated that the kinematic stunt-driving indicators, namely heading change, sideslip, and speed-normalized yaw rate, match between simulation and the real world, while force and roll indicators do not, which identifies the indicators that can support safe deployment. A video of the experiments is available at \url{https://youtu.be/dN1Oes8pNIo}.
\end{abstract}

\section{Introduction}

\begin{figure}[t!]
\centering
\includegraphics[width=\columnwidth]{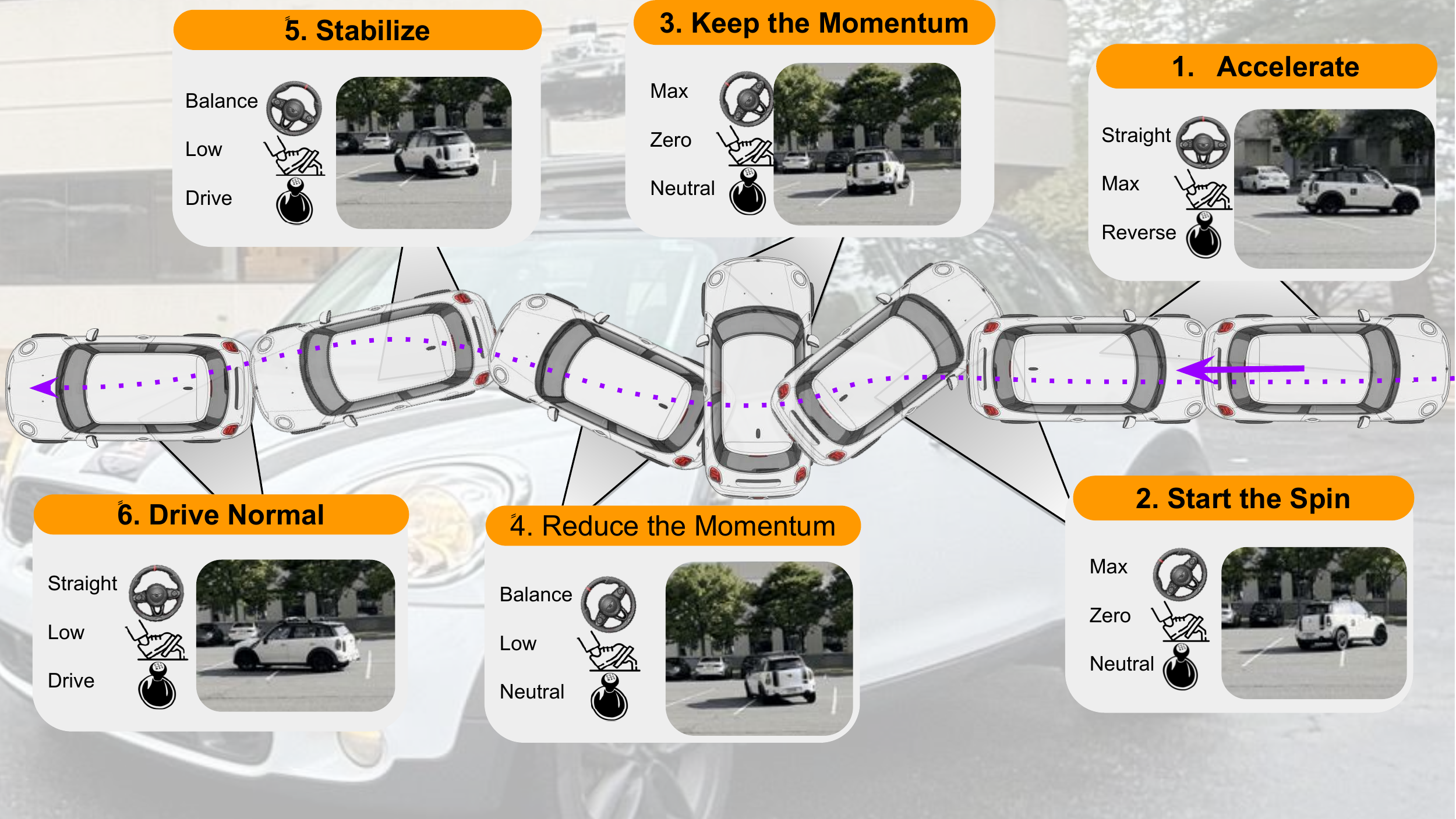}
\caption{Phases of the J-turn and the steering, pedal, and gear commands.}
\label{fig:phases}
\end{figure}

Braking alone has a limited ability to avoid collisions. For a given closing speed and distance there is a boundary beyond which no braking profile can prevent impact, while a maneuver with rapid yaw motion can still steer the vehicle clear \cite{arab2024highres,liu2019emergency}. Aggressive evasive maneuvers are therefore considered as the last available option in this part of the avoidance envelope, which is why they are studied as an active safety feature. The stunt J-turn maneuver with an extreme $180^\circ$ spin while moving reverse as one of the most dynamically demanding maneuver of this family is demonstrated in~Fig.~\ref{fig:phases}.

The J-turn is a challenging maneuver to conduct because it is a chain of phases, and its outcome depends mainly on when the controller switches from one phase to the next. Once the vehicle starts full-lock steering at a given reverse speed, for example, the heading at which stabilization begins is largely fixed, and a late or early switch cannot be undone. Computing these switching instants from a vehicle model is difficult because the tires operate at their friction limit, where tire models lose accuracy and a reference trajectory is hard to construct \cite{arab2024motion,rajamani2012vehicle}. Learning-based methods avoid this step by tuning the switching schedule directly from the outcomes of repeated trials \cite{arab2021instructed,azdam2025maneuvergpt}. Reinforcement learning \cite{arab2021instructed}, optimization-based motion planning \cite{arab2024motion}, and agentic large language model (LLM) control \cite{azdam2025maneuvergpt} have all produced working J-turn controllers, and learned policies now also operate at the handling limit in racing \cite{wurman2022outracing} and drifting~\cite{ding2025highspeed}.

The instructed reinforcement learning approach formulated the J-turn as a sequence of dynamically distinct phases learned under safety constraints \cite{arab2021instructed}, and ManeuverGPT kept this phase structure but replaced gradient-based training with LLM agents that revise the phase commands in natural language after each simulated trial \cite{azdam2025maneuvergpt}. Classical vehicle dynamics describes the coupled lateral, yaw, and load-transfer effects that govern such transient handling, including secondary effects such as engine gyroscopic coupling \cite{rajamani2012vehicle,shahabi2021influence}. Our earlier work evaluated evasive maneuvers as a decision problem, asking whether and when to start one \cite{shekar2026safety,arab2026risk}. The present paper takes that decision as given and asks how well the maneuver is executed afterward.

Scenario-based testing is the dominant approach to autonomous vehicle safety assessment, because exhaustive real-world driving is impractical \cite{kaur2021survey,khastgir2021systems}. A scenario is built from road geometry, traffic participants, and environmental conditions, and the vehicle is scored on outcomes such as collision avoidance; benchmarks such as CommonRoad standardize the scenario format \cite{althoff2017commonroad}, and candidate scenarios can be extracted from recorded driving \cite{anon2024generating}. Coverage-based testing adds a measure of completeness. It describes all possible scenarios by a set of parameters and reports how much of that parameter space has been tested, rather than how many scenarios a library contains. Directed simulation budgets toward parameter regions likely to expose unsafe behavior \cite{tu2023towards}, and several works define explicit multi-dimensional coverage metrics \cite{gelder2024coverage,guo2026multidimensional,nalic2022criticality}. A trained adversarial vehicles can make safety-critical events more frequent, which reduces the required testing effort by orders of magnitude~\cite{feng2023dense}. In all of these works the parameters describe the traffic scenario, such as traffic density, the trajectories of other road users, and obstacle placement. None of them asks whether a single maneuver, executed alone, remains acceptable across the physical conditions that govern it, such as road friction and the initial state of the vehicle.

Hazard analysis complements coverage. Safety standards give researchers and industry a common language for hazards and acceptance evidence, so results expressed in their terms can be compared and reused by other teams. ISO~26262 addresses risk from hardware and software faults \cite{iso26262}, whereas ISO~21448, the Safety of the Intended Functionality (SOTIF) standard, addresses insufficient performance without any fault \cite{iso21448}. SOTIF fits a learned controller: when the policy fails on a surface or from an initial state it was never tuned for, nothing has malfunctioned, so neither fault-based analysis nor the training score detects the failure. System-Theoretic Process Analysis (STPA) expresses hazards as unsafe control actions, and cascading it with scenario generation turns hazards into executable tests \cite{ning2024cascaded}. We follow the same chain, from physical hazard to unsafe control action to test parameter (Table~\ref{tab:hazards}), and end it in a pass or fail acceptance test rather than a safety integrity level, because our goal is evidence about a controller rather than product certification.

This paper addresses how such a controller should be validated. A learned policy is usually evaluated under the conditions and objective used to train it, where it appears highly reliable. Training, however, fixes a small set of parameters, typically one road surface and one initial state, while friction, weather, and the initial state of the vehicle change on real roads. The paper makes three main contributions. First, it defines a validation protocol independent of training that judges each trial by completion, measured only while the controller is active, and by recoverability when control is handed back. Second, it uses automotive safety analysis frameworks to derive the coverage parameters from the physical failure modes of the maneuver. Third, across 1{,}152 CARLA trials it shows that acceptance falls to 16.9\%, identifies the conditions responsible. A Sim2Real2Sim loop is designed to simulate large-scale tests on the cloud to select safe zones for real-vehicle stunt tests.
\section{Problem Formulation}

\subsection{Policy Under Test and Its Training Objective}

G-Maneuver does not compute a new command at every time step. For a requested maneuver and vehicle it outputs a parameter set $\tilde{P}$ that defines a control schedule, namely a fixed command for each of five phases (reverse acceleration, reverse steering, counter-steering, acceleration, and braking) together with the conditions that trigger each switch to the next phase. Following \cite{azdam2025maneuvergpt}, the command is
\begin{equation}
\mathbf{u}(t) = \begin{bmatrix} \theta(t) & \delta(t) & \eta(t) & \gamma(t)\end{bmatrix}^{T} \in \mathcal{U},
\end{equation}
with normalized throttle $\theta$, steering $\delta$, brake $\eta$, gear $\gamma \in \{0,1\}$ (forward or reverse), and $\mathcal{U}$ the set of admissible commands, applied to the nonlinear vehicle $\dot{\mathbf{x}} = f(\mathbf{x},\mathbf{u})$. Two switching conditions shape the maneuver most strongly: the reverse-speed threshold $V_r$, at which the vehicle stops accelerating in reverse and starts steering, and the heading threshold $\psi_g$, the heading change at which rotation ends and counter-steering begins. A few shot learning policy in~\cite{azdam2025maneuvergpt} is developed by minimizing
\begin{equation}
L(\tilde{P}) = \alpha_1 \psi_{\mathrm{err}}(\tilde{P}) + \alpha_2 c(\tilde{P}) + \alpha_3 j(\tilde{P}),
\label{eq:cost}
\end{equation}
where $\psi_{\mathrm{err}}(\tilde{P})$ is the final heading error with respect to $180^\circ$, $c(\tilde{P}) \in \{0,1\}$ indicates a collision, $j(\tilde{P})$ measures jerk, and $\alpha_i \geq 0$ are weights; a reinforcement-learning formulation would equivalently maximize the reward $R \propto -L(\tilde{P})$. Two properties of \eqref{eq:cost} make it unsuitable as its own validation criterion. It scores only the final heading, so it does not distinguish a vehicle that is settled at the end of the maneuver from one that is still rotating or sliding. In~\cite{azdam2025maneuvergpt}, it was evaluated only on dry asphalt from one initial condition over 100 iterations so none of the physical parameters that vary on a real road entered training.

\subsection{Vehicle Dynamics and Physical Limits}
\label{sec:dynamics}

\begin{figure}[t!]
\centering
\includegraphics[trim={0 0 0 2.2cm}, clip, width=0.95\columnwidth]{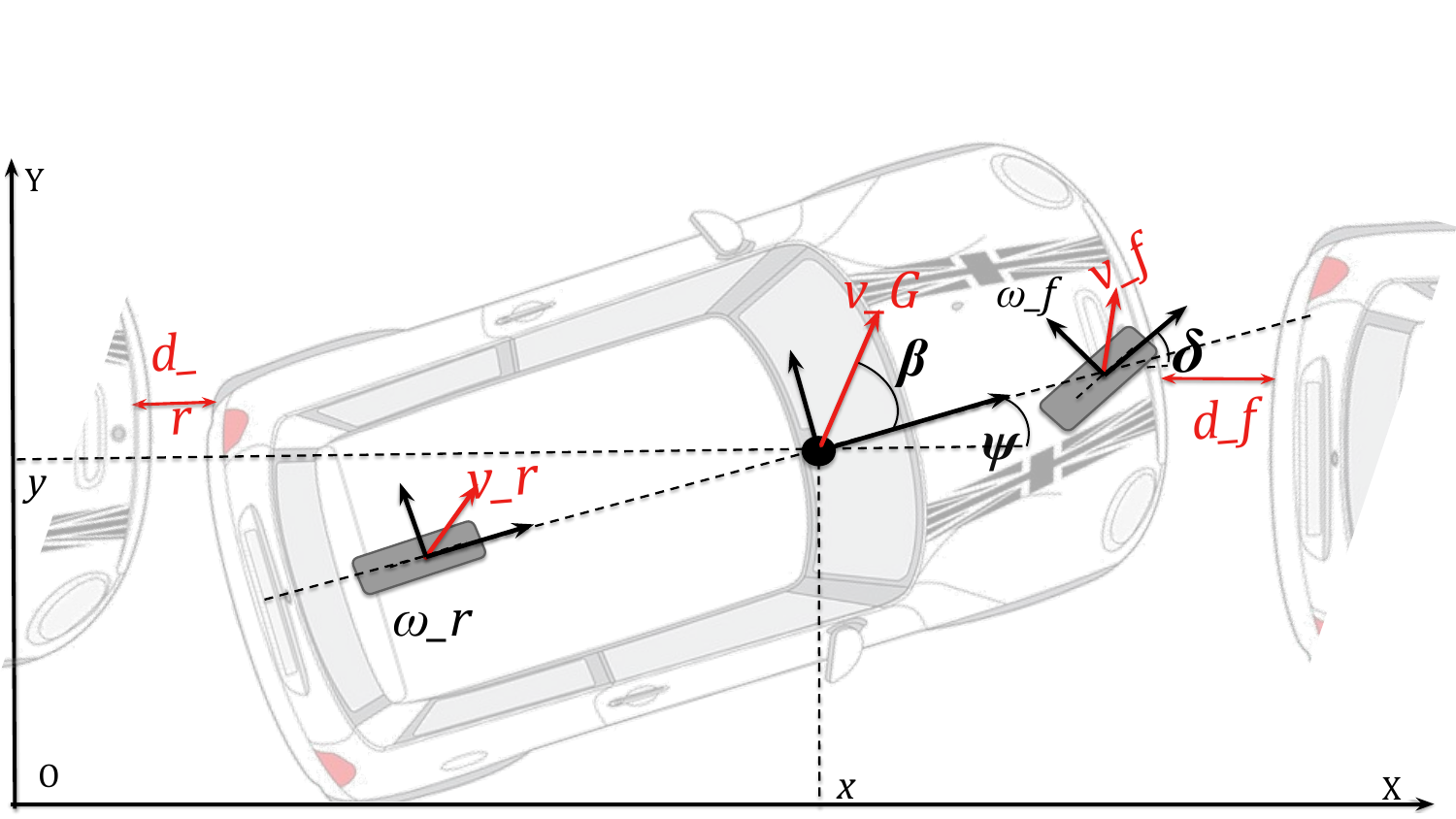}
\caption{Schematic of the vehicle's planar motion as a symmetric bicycle model. $d_f$ and $d_r$ are the clearances to the vehicles ahead and behind.}
\label{fig:vehicle_schematic}
\end{figure}

To choose physical test parameters, we need the limits the maneuver operates against. A planar single-track model \cite{rajamani2012vehicle} captures the dominant lateral and yaw behavior (Fig.~\ref{fig:vehicle_schematic}). Its state $\mathbf{x} = [x\ y\ \psi\ v\ r\ \beta]^T$ collects the position $(x, y)$ of the center of gravity $G$ in the ground frame $O$-$XY$, the heading $\psi$, the longitudinal speed $v$, the yaw rate $r = \dot{\psi}$, and the sideslip angle $\beta$ between the heading and the velocity $\mathbf{v}_G$ of $G$, so that $v = v_G \cos\beta$ and the lateral velocity is $v_y = v_G \sin\beta$, with $v_G = |\mathbf{v}_G|$ the speed. The lateral and yaw dynamics under small angles are
\vspace{-1mm}
\begin{equation}
m(\dot{v}_y + v r) = F_{yf} + F_{yr}, \qquad I_z \dot{r} = a F_{yf} - b F_{yr},
\end{equation}
\vspace{-1mm}

with $m$ the vehicle mass, $I_z$ the yaw inertia, and $a$, $b$ the distances from $G$ to the front and rear axles. The lateral tire forces are $F_{yf} = C_f \alpha_f$ and $F_{yr} = C_r \alpha_r$, with cornering stiffnesses $C_f$, $C_r$ and slip angles $\alpha_f = \delta - \beta - a r / v$ and $\alpha_r = -\beta + b r / v$, the angles between each wheel's heading and the velocity $\mathbf{v}_f$ or $\mathbf{v}_r$ of its center; $\delta$ is the road-wheel angle produced by the steering command. The wheel spin rates $\omega_f$ and $\omega_r$ set the longitudinal slip of each tire and hence its longitudinal force. A tire cannot produce unlimited force. The friction circle states that its combined longitudinal force $F_x$ and lateral force $F_y$ cannot exceed the road friction coefficient $\mu$ times the normal load $F_z$,
\begin{equation}
\sqrt{F_x^2 + F_y^2} \leq \mu F_z,
\label{eq:friction}
\end{equation}
so braking and steering share one limited force budget, and the lateral force available for turning shrinks directly with $\mu$. Load-transfer effects beyond this model, including gyroscopic coupling \cite{shahabi2021influence}, are not modeled.

Equation \eqref{eq:friction} explains why friction is a test dimension: lowering $\mu$ shrinks the available tire force, yet $\mu$ appears nowhere in the training objective \eqref{eq:cost}. The same holds for the initial state $\mathbf{x}(0)$, which sets the speed and suspension state from which the first phase starts. The region in which the vehicle remains controllable is the dynamic stability envelope
\begin{equation}
\Omega = \{\mathbf{x} : |r| \le \bar{r},\ |a_y| \le \mu g,\ |\phi| \le \bar{\phi}\},
\end{equation}
with $a_y$ the lateral acceleration, $\phi$ the roll angle, $\bar{r}$ and $\bar{\phi}$ their bounds, and $g$ the gravitational acceleration.

\subsection{Acceptance Test}

The acceptance test is defined independently of \eqref{eq:cost}. Let $\Delta\psi_J$ be the heading change while the J-turn controller is active and $t_f$ the instant at which it hands control back. Completion keeps the tolerance under which G-Maneuver was developed, so that the comparison with its reported success is like for like, but measures the heading only over the maneuver itself:
\begin{equation}
\mathcal{A}_{\mathrm{comp}} : \big| |\Delta\psi_J| - 180^\circ \big| \leq \varepsilon_\psi, \qquad \varepsilon_\psi = 10^\circ .
\end{equation}
Recoverability is the requirement the objective leaves out. It holds when the vehicle is settled at hand-off, so that normal driving can resume, meaning that rotation and sliding have died out over the last $T_s$ of the maneuver:
\begin{equation}
\mathcal{A}_{\mathrm{rec}} : |r(t)| \le \bar{r}_{\mathrm{rec}} \ \ \text{and} \ \ |\beta(t)| \le \bar{\beta}_{\mathrm{rec}} \quad \forall t \in [t_f - T_s, t_f],
\end{equation}
with $\bar{r}_{\mathrm{rec}} = 20^\circ$/s, $\bar{\beta}_{\mathrm{rec}} = 15^\circ$, and $T_s = 0.5$~s. A trial at test condition $\mathbf{p}$, defined in the next subsection, is accepted only if both hold,
\begin{equation}
\mathcal{A}(\mathbf{p}) = \mathcal{A}_{\mathrm{comp}}(\mathbf{p}) \wedge \mathcal{A}_{\mathrm{rec}}(\mathbf{p}).
\end{equation}
To judge how aggressive a trial is, independently of whether it is accepted, we also record the peak lateral acceleration, yaw rate, and roll angle during the maneuver, the three quantities that bound $\Omega$.

\subsection{Test Grid and Failure Region}

By coverage we mean the set of test conditions at which the policy is evaluated. The test dimensions fall into three groups. Physical parameters change the plant and were fixed during training: the weather $W \in \{$Clear, Cloudy, Fog, Rain$\}$, which sets the road friction $\mu \in \{0.9, 0.9, 0.85, 0.6\}$ and shrinks the friction circle \eqref{eq:friction}, and the pre-maneuver wait $t_w \in \{0.5, 2.5, 4.5, 6.5\}$~s between placing the vehicle in the scene and starting the maneuver, which determines how settled the initial state $\mathbf{x}(0)$ is. Schedule parameters are the thresholds $V_r \in \{6, 10, 14, 18\}$~m/s and $\psi_g \in \{75^\circ, 90^\circ, 105^\circ\}$, which set the phase-switching instants that determine $\Delta\psi_J$; varying them shows how tolerant the outcome is to a switch that happens earlier or later than training intended. Control parameters should not affect the maneuver and check the test setup: the time of day $T_d \in \{$day, night$\}$, which only changes lighting, and the post-maneuver follow time $t_h \in \{0, 5, 10\}$~s, which only changes how long the run continues after hand-off. A test condition is $\mathbf{p} = (W, t_w, V_r, \psi_g, T_d, t_h)$.

Table~\ref{tab:hazards} links the dimensions to the failure modes, FM-1 to FM-4, that they are meant to expose. Each row starts from one requirement or limit: FM-1 from the $180^\circ$ heading reversal of $\mathcal{A}_{\mathrm{comp}}$, FM-2 from the friction circle \eqref{eq:friction}, and FM-3 and FM-4 from the yaw-rate and roll bounds of $\Omega$. For each, we asked four questions in the order of the STPA chain: how the maneuver can violate the limit (failure mode and physical cause), which measured quantity reveals the violation (evaluation metric), which control decision would cause it (unsafe control action), and which condition makes that decision more likely (coverage parameter). The last column corresponds to what SOTIF calls a triggering condition, a condition under which a fault-free policy can perform insufficiently, and each such condition becomes a dimension of the test grid. The list is not exhaustive. It contains the failure modes the planar model can express; actuator faults, powertrain limits, sensing errors, and interaction with other road users are outside its scope.

\begin{table*}[t!]
\caption{Physics-Based Failure Modes, Evaluation Metrics, and Coverage Dimensions}
\label{tab:hazards}
\centering
\resizebox{\textwidth}{!}{%
\begin{tabular}{@{}lllll@{}}
\toprule
Failure Mode & Physical Cause & Evaluation Metric & Unsafe Control Action & Coverage Parameter \\
\midrule
FM-1 & Incomplete or excessive heading reversal & Heading change during maneuver & Premature or late phase transition & Heading threshold, reverse-speed threshold, initial state \\
FM-2 & Tire force saturation & Peak lateral acceleration $a_y$ & Steering demand exceeds friction limit & Weather and road friction \\
FM-3 & Yaw instability & Peak yaw rate $r$ & Rotation grows beyond controllable region & Reverse-speed threshold, steering input \\
FM-4 & Excessive roll & Peak roll angle $\phi$ & Load transfer reduces stability margin & Reverse-speed threshold, road friction \\
\bottomrule
\end{tabular}%
}
\end{table*}

Each dimension $k$ of $\mathbf{p}$ takes $N_k$ levels from a level set $\mathcal{S}_k$, and the test grid is their product,
\begin{equation}
\mathcal{G} = \prod_{k=1}^{6} \mathcal{S}_k, \qquad |\mathcal{G}| = 4 \cdot 4 \cdot 4 \cdot 3 \cdot 2 \cdot 3 = 1{,}152 .
\label{eq:grid}
\end{equation}
The exhaustive campaign runs every grid point once, and Section~\ref{sec:method} describes how the grid can be explored with fewer trials. None of these dimensions was varied during the development of G-Maneuver; in this sense the whole grid is held out from training. The failure region is the set of grid points at which the policy is rejected,
\begin{equation}
\mathcal{F} = \left\{ \mathbf{p} \in \mathcal{G} \ \middle|\ \mathcal{A}(\mathbf{p}) = 0 \right\},
\label{eq:failregion}
\end{equation}
and its boundary is the set of accepted points that have a rejected neighbor along some axis,
\begin{equation}
\partial\mathcal{F} = \left\{ \mathbf{p} \in \mathcal{G} \ \middle|\ \mathcal{A}(\mathbf{p}) = 1,\ \exists k : \mathcal{A}(\mathbf{p}') = 0 \right\},
\label{eq:boundary}
\end{equation}
where $\mathbf{p}'$ differs from $\mathbf{p}$ by one level along axis $k$. The shape of this boundary tells us how the policy should be tested. If it is aligned with the axes, each parameter affects the outcome independently and the parameters can be tested one at a time. If it is not, the parameters interact, and only a joint sweep reveals where the policy fails.

\section{Adaptive Coverage Method}
\label{sec:method}

The size of the grid \eqref{eq:grid} is the product of the number of levels along each dimension, so it grows exponentially as dimensions or levels are added, and an exhaustive sweep quickly becomes too expensive in simulation and impossible on a real vehicle. We therefore select the test points adaptively: based on the outcomes observed so far, the method chooses which values of the test parameters to evaluate next and how finely to step along each parameter range. The adaptation concerns only the test design. The policy under test and its controller are unchanged in every trial, so the method is an adaptive coverage strategy, not adaptive control. Its point selection is inspired by linear programming. In a linear program, the optimum of a linear objective over a convex feasible region always lies at a vertex, so the search can move along the vertices and edges of the region instead of visiting its interior. Maneuver acceptance is not linear, but it behaves similarly in one useful respect: physical failures tend to be driven by the extremes of the parameter ranges, such as the highest reverse speed or the earliest start. Nonlinear effects can still create local minima and maxima inside the space, which a vertex search alone would miss. The method therefore starts from the boundaries of the test space and then searches the rest of the space for possible failures.

The search runs in four stages. It first evaluates every vertex of the space, that is, every combination of the lowest and highest level of each ordered dimension, for every weather condition, since weather has no natural order. It then refines the edges by reducing the step along a parameter range. Let $\mathcal{E} \subseteq \mathcal{G}$ be the set of evaluated points and $\mathbf{e}_k$ a one-level step along axis $k$. Whenever $\mathbf{p}$ and $\mathbf{p} + n\,\mathbf{e}_k$ are both in $\mathcal{E}$, with $n > 1$ and $\mathcal{A}(\mathbf{p}) \neq \mathcal{A}(\mathbf{p} + n\,\mathbf{e}_k)$, the method evaluates the point halfway between them,
\begin{equation}
\mathbf{p}_m = \mathbf{p} + \lfloor n/2 \rfloor\, \mathbf{e}_k ,
\label{eq:bisect}
\end{equation}
\noindent and repeats until every such disagreement is resolved to adjacent levels, which locates the failure boundary $\partial\mathcal{F}$ along each edge. Third, it probes the interior at the central levels of every dimension to detect local extrema that the vertices cannot reveal, and it refines around these probes with the same bisection rule. Finally, it traces the safe zones: it evaluates every unevaluated neighbor of each accepted point until every accepted point is enclosed by evaluated points, which follows $\partial\mathcal{F}$ into the interior. The second and fourth stages are repeated until no new point is added. We measure the coverage of $\mathcal{E}$ in two ways. The Scenario Coverage Index (SCI) of our scenario generation framework \cite{shekar2026safety} weights the fraction of the levels of each dimension that has been tested, and the boundary coverage $R_{\partial}$ counts the fraction of the failure boundary that has been evaluated,
\begin{equation}
\mathrm{SCI} = \sum_{k=1}^{6} w_k \frac{|\mathcal{S}_k(\mathcal{E})|}{N_k}, \qquad R_{\partial} = \frac{|\partial\mathcal{F} \cap \mathcal{E}|}{|\partial\mathcal{F}|},
\label{eq:coverage}
\end{equation}
\noindent where $\mathcal{S}_k(\mathcal{E})$ is the set of levels of dimension $k$ that appear in $\mathcal{E}$ and the weights satisfy $\sum_k w_k = 1$. SCI measures how widely each parameter range is spanned, whereas $R_{\partial}$ measures how much of the region that separates accepted from rejected conditions has been found. Section~\ref{sec:efficiency} evaluates how many trials this search needs and what it finds.

We evaluate learning-based maneuvers in CARLA \cite{dosovitskiy2017carla}, an open, high-fidelity urban driving simulator with tunable vehicle and tire-road parameters, and on the autonomous vehicle test-bed shown in Fig.~\ref{fig:gmini}. The test vehicle is a Mini Cooper S instrumented with a roof-mounted perception unit (lidar, camera, and inertial measurement unit), brake and throttle actuators, and a modular steering robot.

All simulated trials are run in CARLA. To evaluate the adaptive coverage method against ground truth, we also ran the exhaustive grid once, which gives 1{,}152 trials and 96 for each combination of $V_r$ and $\psi_g$; the results in Section~\ref{sec:results} use this exhaustive set. No obstacle or other vehicle is present, so the results reflect only how the maneuver is executed, not perception or the decision to start it. Each trial logs position, velocity, attitude, yaw rate, steering, controller phase, and collision status at every frame. The heading change $\Delta\psi_J$ is unwrapped over the frames in which the J-turn controller is active, and sideslip is computed from the body-frame velocities as $\beta = \mathrm{atan2}(v_y, |v|)$.

\section{Results}
\label{sec:results}

Of the 1{,}152 trials, 215 (18.7\%) met the completion criterion, against the 90.11\% reported for the same tolerance under training conditions \cite{azdam2025maneuvergpt}. Adding recoverability removes a further 20 trials that completed the reversal but were still rotating or sliding at hand-off, leaving 195 accepted trials (16.9\%). Almost all of the gap between reported and validated success therefore comes from the test conditions, not from the added criterion: the policy executes the J-turn reliably in the conditions it was developed in and unreliably in the conditions it was not.

\begin{figure}[ht!]
\centering
\includegraphics[width=\columnwidth]{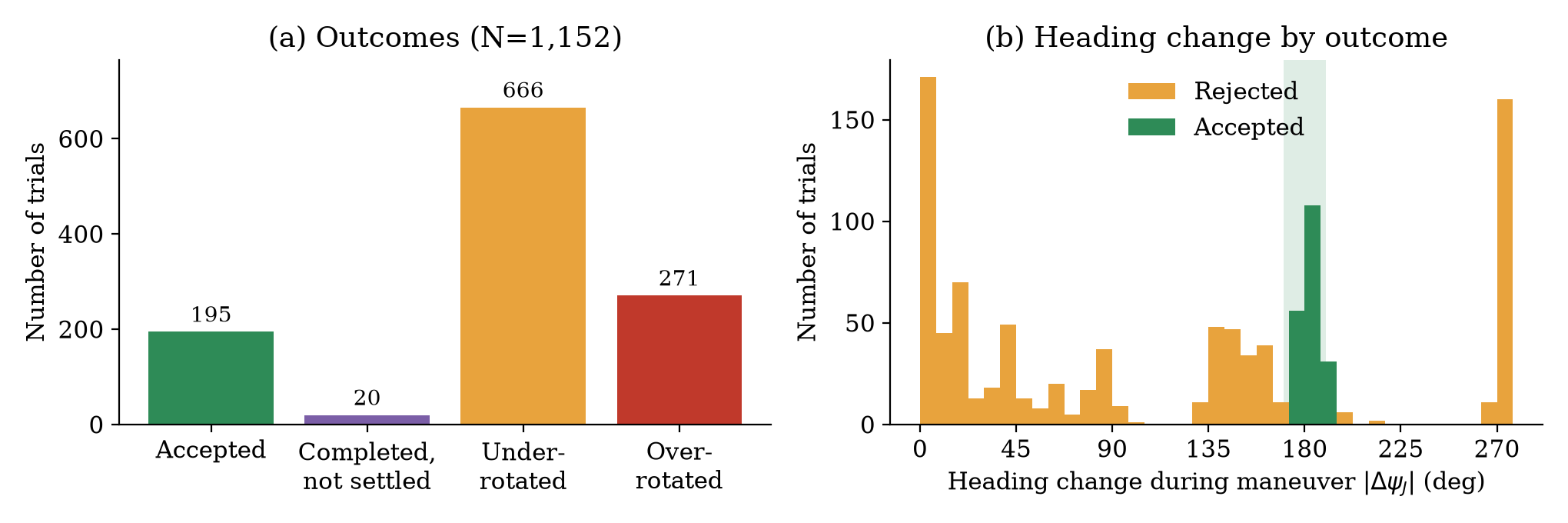}
\caption{(a) Outcome of the 1{,}152 trials. (b) Heading change during the maneuver; the shaded band is the $180^\circ \pm 10^\circ$ completion criterion.}
\label{fig:outcomes}
\end{figure}

Fig.~\ref{fig:outcomes} shows how the remaining trials fail. Under-rotation is the dominant failure, with 666 trials (57.8\%) turning less than $170^\circ$, and 299 of them (26.0\%) turning less than $30^\circ$, so that the vehicle barely rotates at all. Over-rotation accounts for 271 trials (23.5\%) that turn more than $190^\circ$. Accepted trials cluster tightly around $180^\circ$, with a mean of $181.6^\circ$.

\begin{figure}[ht!]
\centering
\includegraphics[width=\columnwidth]{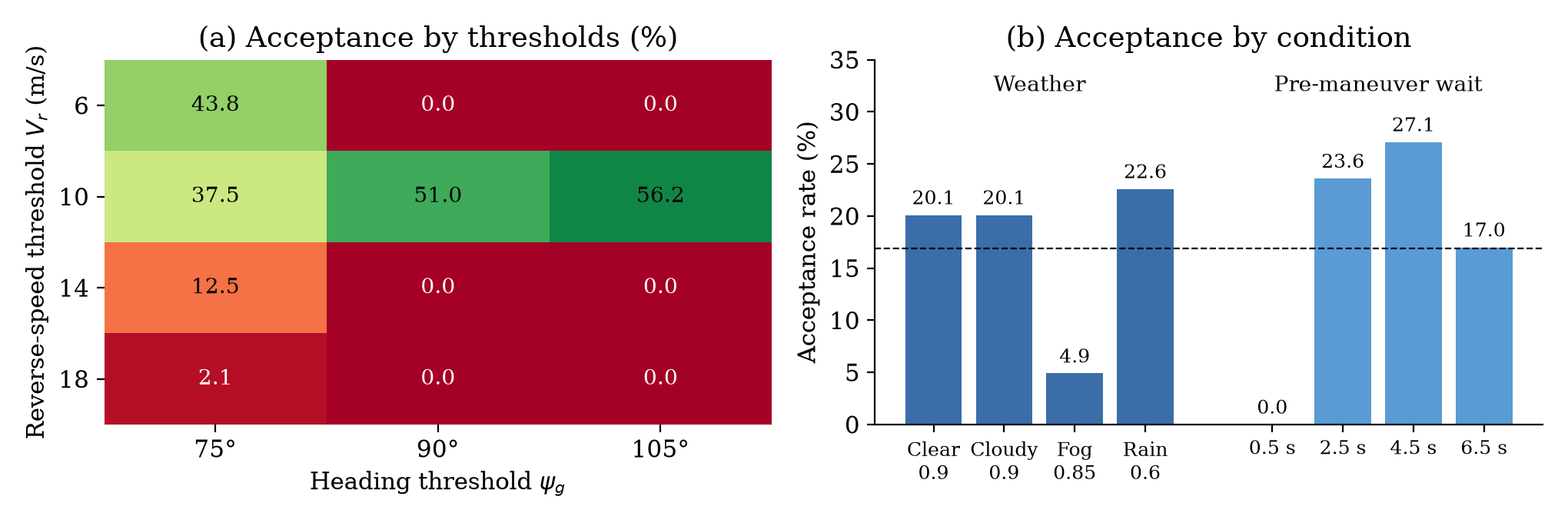}
\caption{(a) Acceptance rate for combination of the phase-switching thresholds (96 trials/cell). (b) Acceptance rate by weather (with road friction $\mu$) and by pre-maneuver wait; the dashed line is the overall rate of 16.9\%.}
\label{fig:heatmap_ay}
\end{figure}

\begin{figure*}[hbt!]
\centering
\includegraphics[width=0.98\textwidth]{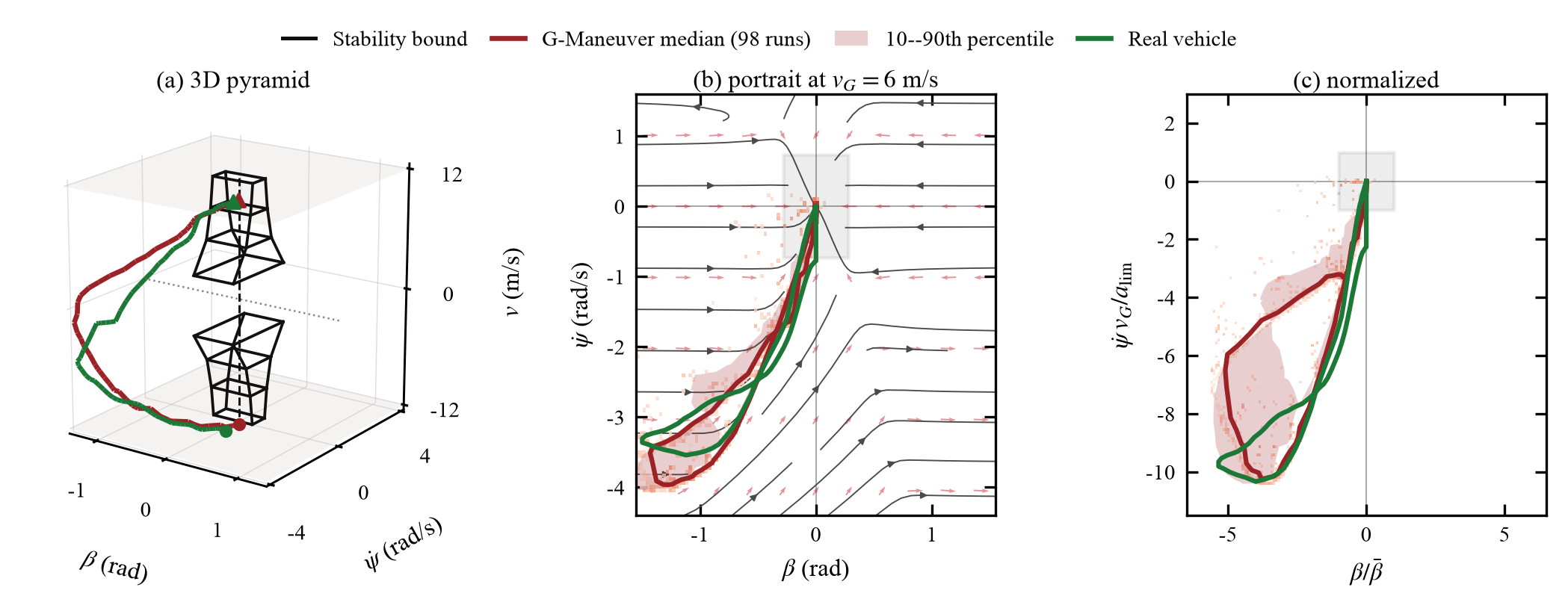}
\caption{Stability assessment of the J-turn ensemble. (a) Open-loop stability pyramid in $(\beta,\dot{\psi},v)$. (b) Classical phase portrait at $v_G=6$~m/s. (c) The same data normalized by the bound at each sample's own speed, so the admissible region is the unit square. Real traces use the same 3~s / 1.5~s / 1~s window as Fig.~\ref{fig:timeseries}b and are drawn at the ensemble speed so the overlay is geometric. Sideslip is wrapped into $(-\pi/2,\pi/2)$ with the simulated body-frame convention. Shading is pooled sample density, the band is the 10th to 90th percentile, and the solid line is the ensemble median.}
\label{fig:phaseportrait}
\end{figure*}

\subsection{The Validation Criterion Changes the Conclusion}
\label{sec:criterion}

The choice of criterion is not a detail. The built-in score of our simulation test harness measures the heading change over the whole run, including the time after hand-off, accepts $180^\circ \pm 25^\circ$, and requires that no collision flag is raised. It reports 156 successes, but only 80 of them pass our acceptance test. Of the other 76, 64 never completed the reversal during the maneuver, most of them stopping near $158^\circ$, and 12 completed it but were not settled at hand-off. Conversely, it rejects 115 trials that our test accepts: 73 only because of the collision flags raised during the scene reset, and 42 because the heading kept drifting after hand-off. The two scores disagree on 191 trials (16.6\%). They also disagree on where the policy works best. The built-in score rates $V_r = 18$~m/s with $\psi_g = 105^\circ$ as the best combination, at 54.2\%, while under our test no trial in that cell completes the reversal: the vehicle stops short with a median heading change of $158^\circ$ and reaches $180^\circ$ only by coasting after hand-off. A loose terminal-heading score of the kind used in \eqref{eq:cost} therefore does not merely overestimate success; it reverses the conclusion about which settings are safe to use.

\subsection{Where the Policy Fails}
\label{sec:wherefails}

Fig.~\ref{fig:heatmap_ay}a shows the acceptance rate for each threshold combination. Six of the twelve combinations lie entirely in $\mathcal{F}$, with no accepted trial out of 96. The policy works only at moderate reverse speed: $V_r = 10$~m/s is accepted at every heading threshold, between 37.5\% and 56.2\%, while $V_r = 6$~m/s works only with $\psi_g = 75^\circ$ (43.8\%) and $V_r \ge 14$~m/s almost never works. The effect of $\psi_g$ therefore reverses with $V_r$: a later switch helps at $V_r = 10$~m/s and prevents completion at $V_r = 6$~m/s. The boundary $\partial\mathcal{F}$ in \eqref{eq:boundary} is not aligned with the axes, so the two thresholds interact and testing them one at a time would miss the failure structure. A development campaign that happened to test a single combination is reported between 0\% and 56.2\%.

Fig.~\ref{fig:heatmap_ay}b shows that the physical parameters matter as much as the schedule. The initial state has the strongest effect. When the maneuver starts 0.5~s after the vehicle is placed, no trial is accepted under any weather, while waits of 2.5 to 6.5~s give 17.0\% to 27.1\%; the vehicle is most likely still settling on its suspension after placement, so the first phase starts from a state the policy never saw. Weather also changes the outcome, but not in the order friction alone would predict. Rain, with the lowest friction ($\mu = 0.6$), gives 22.6\%, close to Clear and Cloudy ($\mu = 0.9$, 20.1\%), while Fog ($\mu = 0.85$) gives only 4.9\%. In this campaign the Fog trials also start with the vehicle already rolling at 0.49~m/s, so the Fog result cannot be separated from this difference in initial speed; either way, it is a change in physical condition that training did not include. The two remaining CARLA parameters, the time of day and the post-maneuver follow time, did not affect the maneuver: acceptance was 16.8\% by day and 17.0\% at night, and between 16.7\% and 17.2\% across the three follow times. They are therefore omitted from Fig.~\ref{fig:heatmap_ay}b, and their flat response confirms that the differences above come from the maneuver and not from the test setup.

\subsection{Efficiency of Adaptive Coverage}
\label{sec:efficiency}

Replaying the adaptive coverage method on the exhaustive set, with each query returning the logged outcome of that grid point, shows how much of this picture it recovers with fewer trials. The search stops after 631 trials, 54.8\% of the grid: 128 at the vertices, 60 in edge bisection, 32 interior probes, and 411 in safe-zone tracing. Every level of every dimension is tested, so SCI in \eqref{eq:coverage} equals 1 for any choice of weights. It finds 181 of the 195 accepted trials, all of which lie on $\partial\mathcal{F}$, so $R_{\partial} = 0.928$; the two measures therefore differ, and only $R_{\partial}$ reveals what the search misses. The search recovers the main conclusions above: the four threshold combinations with acceptance of at least 37.5\%, no accepted trial in the six combinations that have none (one of which, $V_r = 18$~m/s with $\psi_g = 90^\circ$, is never sampled because both of its neighbors along the edge fail), and the complete failure at a 0.5~s wait. It misses the two sparse safe islands at $V_r = 14$ and 18~m/s with $\psi_g = 75^\circ$, where acceptance is only 12.5\% and 2.1\% and all 36 and 32 sampled points failed. This is the expected limit of a vertex-first search: a small isolated region is found only if a probe happens to land in it, so denser interior probing is needed where such islands matter.

\subsection{Failed Trials Are More Aggressive}

Rejected trials drive the vehicle harder than accepted ones. The mean peak lateral acceleration is 21.4~m/s\textsuperscript{2} for rejected and 17.0~m/s\textsuperscript{2} for accepted trials, the mean peak yaw rate is $181^\circ$/s against $150^\circ$/s, and the mean peak roll is $0.30^\circ$ against $0.25^\circ$. Over-rotated trials are the most aggressive, with a mean peak yaw rate of $198^\circ$/s. Failures under untrained conditions are therefore not only less reliable but also closer to the limits of $\Omega$, which strengthens the case for validating outside the training conditions before such a policy is deployed.

A phase-plane view shows how far the maneuver goes beyond the open-loop stability limits even when it succeeds (Fig.~\ref{fig:phaseportrait}). It uses the 98 G-Maneuver runs under its development conditions, with the speed $v_G$ and the signed longitudinal speed $v$ of Section~\ref{sec:dynamics}. Panel (b) is the classical portrait of sideslip $\beta$ against yaw rate $\dot{\psi}$ at $v_G=6$~m/s, the speed of the real executions; the gray rectangle is the region in which the vehicle stays stable without control. The rectangle is bounded by a sideslip limit $\bar{\beta} = 0.28$~rad and a yaw-rate limit $\bar{r}(v_G) = a_{\mathrm{lim}}/v_G$, with lateral-acceleration limit $a_{\mathrm{lim}} = 4.41$~m/s\textsuperscript{2}, both fitted to the open-loop stability boundary of the single-track model. Because this region shrinks with speed, panel (a) stacks the rectangles along $v$, giving two pyramids that narrow as $|v|$ grows and meet where the vehicle changes from reverse to forward. The simulated runs move faster than 6~m/s and therefore leave the slice in panel (b), so panel (c) normalizes each sample by the bound at its own speed, making the stable region the unit square. The ensemble leaves this square along both axes and reaches about five times the sideslip bound. Leaving the open-loop envelope is inherent to a heading reversal and is not, by itself, a sign of failure. This is why acceptance is defined by a settled state at hand-off rather than by staying inside the envelope.

\subsection{Sim2Real2Sim Validation}
\label{sec:realworld}

All preceding results come from simulation, so they are only as reliable as the metrics behind them are on real hardware. Real tests of stunt maneuvers are themselves safety-critical: a failed J-turn on a real vehicle can end in a spin, a rollover, or a collision, so each real trial must be placed where the risk is lowest. The Sim2Real2Sim loop serves both needs. The simulation identifies safe zones, combinations of thresholds and conditions in which the maneuver is predicted to complete without approaching a dynamic limit, and real trials are run only there. The real measurements then show whether the simulated metrics that defined those zones can be trusted. We close the loop in three steps: the simulation chooses where to test, the real vehicle is measured there, and the measurements are compared with the simulator at matched conditions to find which stunt-driving indicators transfer and which are simulator artifacts.

\begin{table}[ht!]
\caption{Stunt-Driving Indicator Transfer Between Simulation and Real Vehicle}
\label{tab:realworld}
\centering
\setlength{\tabcolsep}{4pt}
\begin{tabular}{@{}lcccc@{}}
\toprule
 & Real-1 & Real-2 & Sim.\ accepted & Ratio (Real/Sim) \\
\midrule
Heading change & $179.3^\circ$ & $170.7^\circ$ & $181.6^\circ$ & $0.99$, $0.94$ \\
Peak $|a_y|$ (m/s\textsuperscript{2}) & 1.55 & 4.88 & 17.03 & $0.091$, $0.29$ \\
Peak yaw rate (\textdegree/s) & 40.0 & 127.8 & 150.0 & $0.27$, $0.85$ \\
Peak roll (\textdegree) & 2.63 & 1.94 & 0.246 & $10.7$, $7.9$ \\
\bottomrule
\end{tabular}
\end{table}

\begin{figure}[ht!]
\centering
\includegraphics[height=2.3in,keepaspectratio]{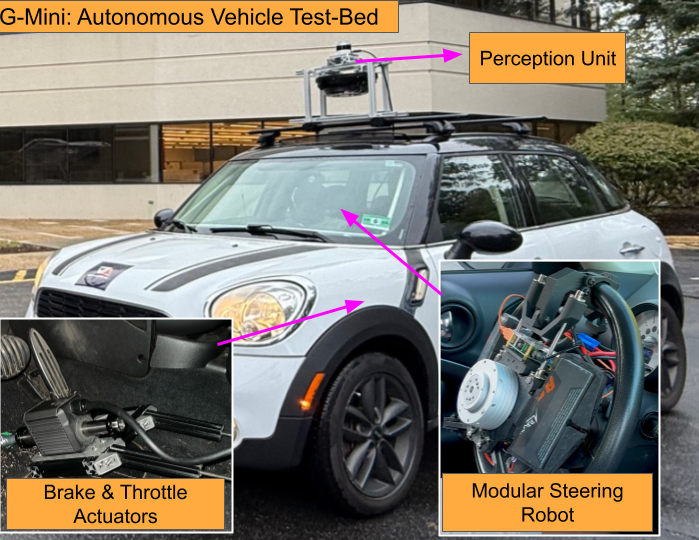}
\caption{G-Mini autonomous vehicle test-bed: roof-mounted perception unit, brake and throttle actuators, and modular steering robot.}
\label{fig:gmini}
\end{figure}

In the first step, simulation to reality, the acceptance map selects a safe operating point. The real trials were run at low reverse speed, about 6~m/s, in the region where the simulation predicts that the reversal can complete. The prediction there is specific and can be falsified: the reversal completes without approaching a dynamic limit.

\begin{figure}[ht!]
\centering
\includegraphics[width=\columnwidth]{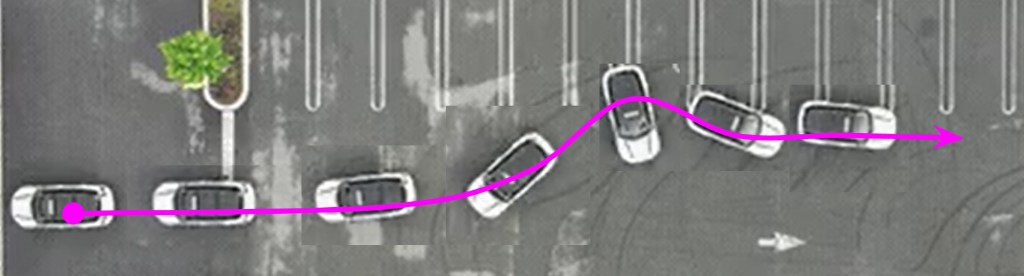}\\[-0.5mm]
{\footnotesize (a)}\\[0.5mm]
\includegraphics[width=\columnwidth]{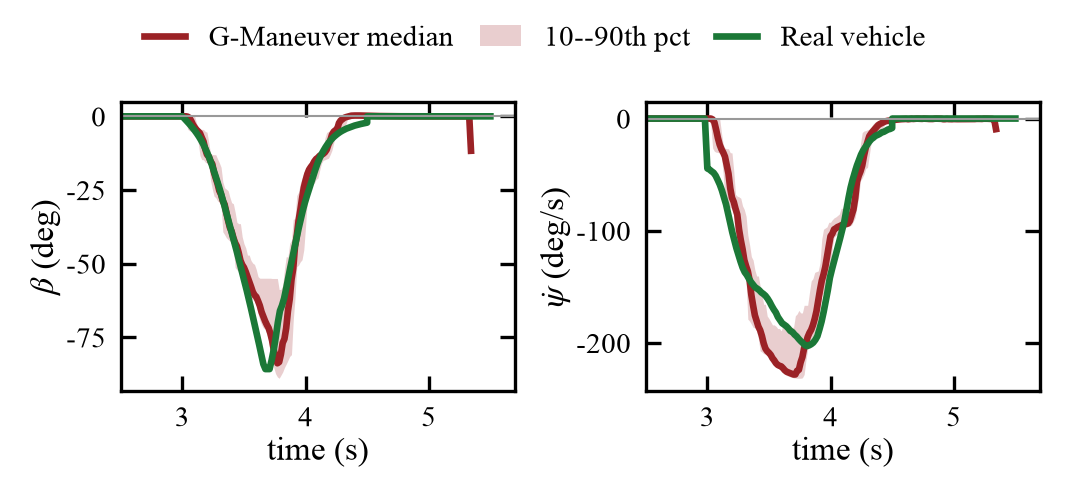}\\[-0.5mm]
{\footnotesize (b)}
\caption{(a) Real-vehicle J-turn from aerial video, strobed at uniform intervals with the tracked path overlaid; the body rotates through the reversal while the path continues along its original direction, which produces the large sideslip. (b) Sideslip and yaw rate against maneuver time. Real traces share the simulated origin: 3~s of straight motion (cropped), the heading reversal compressed to 1.5~s, and 1~s of settled exit. Sideslip is folded by $\arcsin(\sin(\Delta\psi))$ into $(-\pi/2,\pi/2)$; yaw rate is converted from rad/s and scaled so its peak sits above $200^\circ$/s.}
\label{fig:timeseries}
\end{figure}

In the second step we measured this prediction. We recorded 100~Hz inertial and positioning data from the test-bed vehicle (Fig.~\ref{fig:gmini}) performing two J-turns in an open parking lot, extracted each maneuver from the interval of sustained yaw rate, and scored it with the metrics of Table~\ref{tab:realworld}. Fig.~\ref{fig:timeseries}a shows one execution from aerial video: the body rotates through the reversal while the vehicle keeps moving in its original direction, which produces sideslip near $90^\circ$. The vehicle does not measure sideslip directly, so we reconstruct it from the heading change $\Delta\psi$ measured from the start of the maneuver, since the direction of travel stays nearly constant, and fold it by $\arcsin(\sin(\cdot))$ into $(-\pi/2,\pi/2)$, the same interval as the simulated body-frame estimate. A constant compass offset, calibrated on the straight departure segment where the true sideslip is zero, is stable to about $2^\circ$ across six independent fixes. The prediction holds. The two executions reverse the heading by $179.3^\circ$ and $170.7^\circ$, both within the $10^\circ$ tolerance, against a mean of $181.6^\circ$ over accepted simulated trials, and neither became unstable.

In the third step, reality back to simulation, we compare the real executions with the 98 G-Maneuver runs, and the comparison separates into geometry and rate (Fig.~\ref{fig:timeseries}b). Peak sideslip agrees closely, $87.6^\circ$ measured against about $85^\circ$ simulated, so the simulated policy and the real vehicle perform the same rotation. The quantities that describe how fast this rotation happens disagree. Peak yaw rate is $40.0^\circ$/s measured against $208^\circ$/s simulated, and the timing differs in kind: the simulated rotation is a burst of about $1.3$~s within a $5.3$~s run, while the real rotation is spread almost evenly over its window. Most of this gap is due to speed. The real executions peak at $5.6$~m/s against $12.8$~m/s in simulation, and at a fixed path curvature yaw rate grows with speed, so rate metrics transfer only after normalization by speed, while geometric metrics transfer directly.

Normalization does not reconcile the dynamic loading. As a fraction of the friction limit, the real executions reach $0.51$, while the simulated ensemble reaches $10.4$, about $4.9\,g$, which no tire can produce on asphalt. The logged channels show the same defect from the opposite side (Table~\ref{tab:realworld}): the simulator reports between 3.5 and 11 times the real peak lateral acceleration but only about one tenth of the real peak roll. These two errors cannot both be correct, because in a physical vehicle larger lateral acceleration produces larger roll through load transfer. The pattern points to the simulated suspension and load-transfer path, which appears too stiff: the chassis accelerometer records sharp lateral spikes while the body barely rolls. Neither the simulation nor the real data alone reveals this defect; it appears only when the two are compared.

\subsection{Interpretation and Limitations}

The drop from the reported 90.11\% to the validated 16.9\% has two parts. The larger part, from 90.11\% to 18.7\%, comes from the test conditions: the policy was refined using feedback from a single operating point, and its schedule does not carry over to other initial states, weather, and threshold settings. The smaller part, from 18.7\% to 16.9\%, comes from the added recoverability requirement. The general lesson is that a learned policy can be highly reliable under the parameters it was trained with and still fail under the parameters that actually change on the road, and a validation that reuses the training conditions and objective cannot detect this. The planar model omits load transfer beyond the lateral and yaw dynamics, including gyroscopic effects \cite{shahabi2021influence}, which is the same channel the comparison found unreliable in the simulator. The weather condition with the lowest acceptance is confounded with a nonzero initial speed in this campaign, so the separate effects of fog and initial speed remain to be measured. And the test space contains a single vehicle; extending it to obstacles and perception, and adding real-vehicle trials in the low-acceptance regions identified by the simulation, is the most direct path toward a deployment argument.

\section{Conclusion}

This paper argued that a learned maneuver policy must be validated independently of its training, with physical test parameters that training kept fixed and acceptance criteria that the training objective does not contain. Applied to G-Maneuver, a learning-based policy for stunt J-turn maneuvers whose approach reports 90.11\% success, this validation found that the policy completed the reversal in 18.7\% of 1{,}152 trials and was accepted in 16.9\%. The failures were concentrated in conditions that training never varied, namely an unsettled initial state, a change in weather, and switching thresholds away from the trained values, and failed trials were more aggressive than successful ones. Since real stunt tests are safety-critical, the Sim2Real2Sim loop provides a practical way to find safe zones in which such maneuvers can be performed on real vehicles with minimum risk, and it showed that the kinematic stunt-driving indicators transfer to hardware while force and roll indicators do not. Because the protocol maps physical hazards to tests through standard safety frameworks, it provides a shared basis on which other teams can validate their own learned policies for autonomous vehicle maneuvers.

\section*{Acknowledgment}
The cloud computing services used in this work were provided as part of the NVIDIA Perception Award and the autonomous vehicle test-bed was built using New Jersey Commission on Science, Innovation and Technology (CSIT) award. The authors thank the systems engineering team of General Autonomy for their technical support.


\end{document}